\documentclass{ifacconf}

\usepackage{graphicx}      
\usepackage{natbib}        
\usepackage{amsmath}
\usepackage{xcolor}
\usepackage{subcaption} 
\usepackage{booktabs} 
\usepackage{siunitx}  
\usepackage{url}

\begin{document}
\begin{frontmatter}

\title{Strawberry Robotic Operation Interface: An Open-Source Device for Collecting Dexterous Manipulation Data in Robotic Strawberry Farming} 


\author[First]{Linsheng Hou} 
\author[First]{Wenwu Lu} 
\author[Second]{Yanan Wang}
\author[First]{Chen Peng}
\author[First]{Zhenghao Fei}
\thanks{Corresponding author: Zhenghao Fei, e-mail: zfei@zju.edu.cn}
\thanks{Codes and data at: https://github.com/AgRoboticsResearch/SROI}

\address[First]{ZJU-Hangzhou Global Scientific and Technological Innovation Center, Zhejiang University, Hangzhou, 311215, China }
\address[Second]{College of Biosystems Engineering and Food Science, Zhejiang University, Hangzhou, 310058, China }
   
\begin{abstract}                
The strawberry farming is labor-intensive, particularly in tasks requiring dexterous manipulation such as picking occluded strawberries. To address this challenge, we present the Strawberry Robotic Operation Interface (SROI), an open-source device designed for collecting dexterous manipulation data in robotic strawberry farming. The SROI features a handheld unit with a modular end effector, a stereo robotic camera, enabling the easy collection of demonstration data in field environments. A data post-processing pipeline is introduced to extract spatial trajectories and gripper states from the collected data. Additionally, we release an open-source dataset of strawberry picking demonstrations to facilitate research in dexterous robotic manipulation. The SROI represents a step toward automating complex strawberry farming tasks, reducing reliance on manual labor.
\end{abstract}


\end{frontmatter}

\section{Introduction}
The global fresh strawberry market was valued at approximately USD 20.69 billion in 2024 and is projected to grow at a CAGR of 4.40\% year over year accourding to Expert Market Research \cite{Claight}, reaching around USD 30.56 billion by 2034. North America is the largest market for fresh strawberries, primarily driven by the United States. USDA-NASS reported total strawberry production in 2022 of 27.82 million hundredweight, with California producing 89.1\% of them being the leading state in strawberry production (\cite{wade2024review}).  While  the most commonly used growing methods for strawberries are still the matted row system and hill systems due to their simplicity and low cost. The tabletop growing system is becoming a trend in stawberry growing  due to its improved labor efficiency and higher yields to sustainability and space optimization—make it a compelling choice for both commercial growers and small-scale farmers especially meet the demand for organic, high quality and locally-grown produce (\cite{um}). 

While the market and demand for strawberries are growing worldwide, strawberry farming, particularly for tabletop systems, remains a labor-intensive endeavor that involves a variety of tasks requiring substantial human labor. The shortage of agricultural labor is an inevitable trend among the largest strawberry-producing countries, including China, the United States, and Japan. Labor-intensive tasks include stolon removal, leaf removal and thinning, and, most importantly, strawberry picking, which is necessary to harvest the often-clustered fruit. Due to the delicate nature of the plants and fruits, these operations cannot be easily automated. Consequently, robotic manipulation is regarded as one of the few viable methods to reduce reliance on manual labor.

Robotic picking for strawberry farming has not yet reached a conclusive solution, and the field remains an active area of research. The advancements in perception and the development of straightforward planning and control algorithms for robotic arms have made it increasingly feasible to deploy picking robots capable of detecting strawberries and executing a plan-grasp-control paradigm in relatively standardized elevated cultivation environments.

\begin{figure}[t]
\begin{center}
\includegraphics[width=8.4cm]{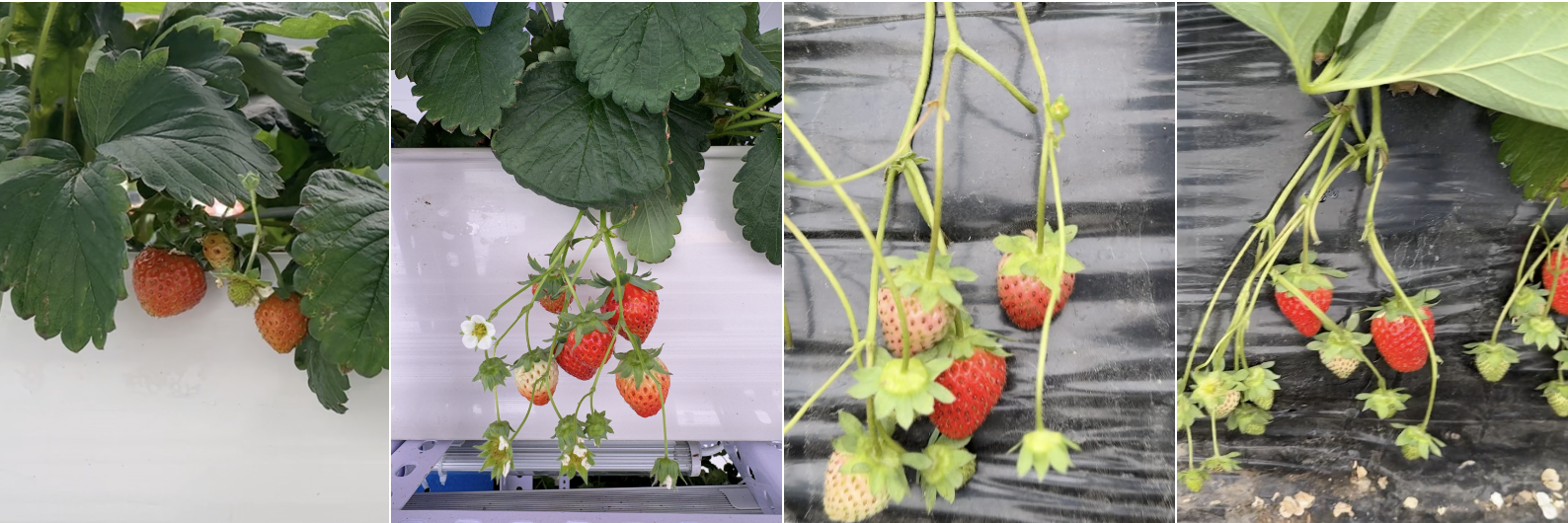}    
\caption{Strawberries that are occluded by stems, leaves, and other strawberries necessitate the use of dexterous robotic manipulation for picking, as they cannot be directly harvested.} 
\label{fig:occluded_strawberry}
\end{center}
\end{figure}

However, the current control methods lack the necessary dexterity, making strawberry picking in unstructured, complex environments with significant obstructions a persisting challenge. This type of dexterous manipulation is also a great challenge in the field of robotics, as it involves dexterous manipulation in unstructured environments where the objects are versatile and deformable biological entities such as in Fig.~\ref{fig:occluded_strawberry} . Such conditions are extremely difficult to model explicitly and solve analytically.

\begin{figure}
\begin{center}
\includegraphics[width=8.4cm]{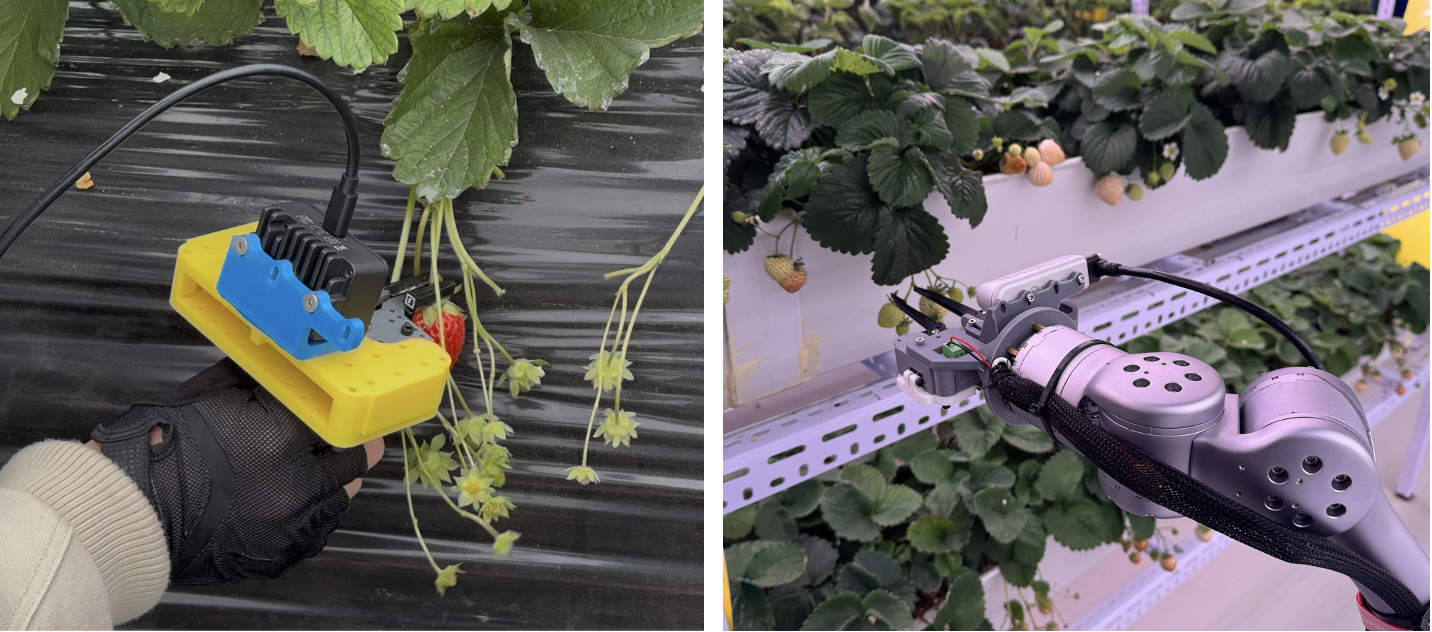}    
\caption{Left: Human demonstration of dexterous occluded strawberry picking using our handheld data collection device. Right: Robot arm strawberry picking using our e-gripper with same apperance as the handheld data collection device} 
\label{fig:sroi_collect_inference}
\end{center}
\end{figure}

One potential approach to overcoming the challenges of dexterous manipulation with deformable objects is through imitation learning. Recent breakthroughs in this area, including (\cite{black2024pi_0}, \cite{aldaco2024aloha}, \cite{zhao2024aloha}, \cite{chi2024diffusionpolicy}, \cite{Zhao2023-hl},  \cite{chi2023diffusionpolicy}), demonstrate the feasibility of employing imitation learning with specific architectures, such as Conditional Variational Autoencoders (\cite{Zhao2023-hl}, \cite{sohn2015learning}), diffusion policy (\cite{chi2023diffusionpolicy}, \cite{ho2020denoising}, or flow matching (\cite{lipman2022flow}, \cite{black2024pi_0} ), to learn manipulation policies capable of controlling robots to perform tasks that were previously impossible or exceedingly difficult to code. These tasks include cloth folding, shoe lacing, plastic bag opening, dish loading, and similar activities that involve the manipulation of deformable objects, dealing with unmodeled external forces, and addressing numerous unseen conditions. We believe these types of imitation learning methods could be crucial in solving various robotic operations in strawberry farming, including the picking of occluded strawberries in clusters.

However, imitation learning requires a substantial amount of human demonstration data, and collecting this data is particularly challenging for field robotics. \cite{chi2024diffusionpolicy}, \cite{aldaco2024aloha}, and \cite{zhao2024aloha} allow a human to operate two twin arms behind the real robot and map the actions to the real arm to collect data, yet this kind of device is difficult to deploy in the field due to its size and lack of mobility.  \cite{cheng2024tv} employs a virtual reality (VR) device to capture the hand movements of human operators and map the hand trajectory to the robot arm's end trajectory. This method of data collection necessitates commercial VR headsets and joysticks, which can be prohibitively expensive for large-scale field data collection, and the commercial VR headsets often lack an open API to access all the required data for teleoperating a robot.

The most relevant work to date is the UMI (Universal Manipulation Interface) data collection device and its variant (\cite{chi2024universal}, \cite{wu2024fast}). UMI utilizes a GoPro camera mounted on an unpowered handheld gripper to capture first-person view videos of the gripper's movements. The gripper's trajectory can be recovered using monocular SLAM. This approach offers several advantages, including being open-source, low-cost, easy to deploy in the field, and suitable for distribution to multiple operators for collaborative data collection. During robot control, a powered gripper with the same camera view will be employed, ensuring that the learned policy remains robot-agnostic. However, the UMI system has several limitations. First, it relies on a GoPro camera, which is not specifically designed for robotic applications. Utilizing such a camera as a robot's primary camera can be expensive and dependent on specific interface hardware, such as the GoPro Media Mod and HDMI capture card, which can be cumbersome for field robotics. Second, the camera provides a monocular view. Although UMI incorporates two side mirrors for implicit stereo vision, no explicit stereo information can be retrieved. This limitation may hinder the learning of operations that have high depth estimation requirements. Third, the end effector of UMI is not designed specifically for strawberry farming operations. While its wide and soft thermoplastic polyurethane (TPU) fingers are suitable for various household tasks, such as cloth folding and cup arrangement, they are not capable of picking clustered strawberries or removing leaves. Thus, we designed SROI as shown in Fig. \ref{fig:sroi_collect_inference}, a specifically tailored device for collecting data on dexterous robotic manipulation in strawberry farming operations. This system includes a portable data collection unit capable of powering the camera, controlling data recording, and temporarily storing the data. Additionally, a post-processing system has been developed to convert the raw recorded data into manipulation trajectories, including the spatial poses of the end effector and the states of the gripper.

The major contributions of this work include the following:
\begin{enumerate}
\item The design of the SROI specifically tailored for collecting dexterous robotic manipulation data related to strawberry farming, particularly for strawberry picking.
\item The design of a low-cost, open-source electric gripper for mounting on the robot arm, with the same camera perspective as the handheld data collection device for seamless application of a learned policy.
\item The collection and open-source release of a strawberry picking manipulation dataset, which includes strawberry picking trajectories in various scenes, along with the corresponding data processing pipeline to estimate the spatial trajectory for each operational demonstration.
\item Evaluation of the accuracy of spatial trajectory estimation derived from the data collected by the SROI.
\end{enumerate}

\section{Materials and Methods}

The SROI is specifically designed for the collection of dexterous robotic manipulation data related to strawberry harvesting. It incorporates several key modifications to the original UMI:

\begin{enumerate}
 \item \textbf{Strawberry Operation End Effector:} A cut-and-hold end effector has been developed for the SROI, making it suitable for tasks such as strawberry picking, leaf removal, and stolon removal. Additionally, the end effector interface is modular designed, allowing for the integration of different types of end effectors.
 
\item \textbf{Low-Cost Robot-Side Electric Gripper:}  A low-cost, open-source electric gripper has been designed for direct mounting on the robot arm. The camera perspective aligns precisely with that of the handheld data collection device, facilitating the seamless application of the learned policy.

\item \textbf{Stereo Robotic Camera:} The GoPro camera has been replaced with a compact stereo camera commonly used in robotic applications, such as the Realsense D435i or OAK-D-SR. This new camera captures stereo data along with IMU information and connects to any computer via a USB cable, eliminating the need for an HDMI capture card.

\item \textbf{Portable Data Collection Unit: } A standalone portable data collection unit has been designed to power the camera and collect demonstration data in a standard format.

\item \textbf{Compact Size:} The overall handheld device has been redesigned to be significantly more compact, enhancing its usability in dense and crowded operational environments, such as those found near strawberry plants.
\end{enumerate}

\begin{figure}
\begin{center}
\includegraphics[width=8.4cm]{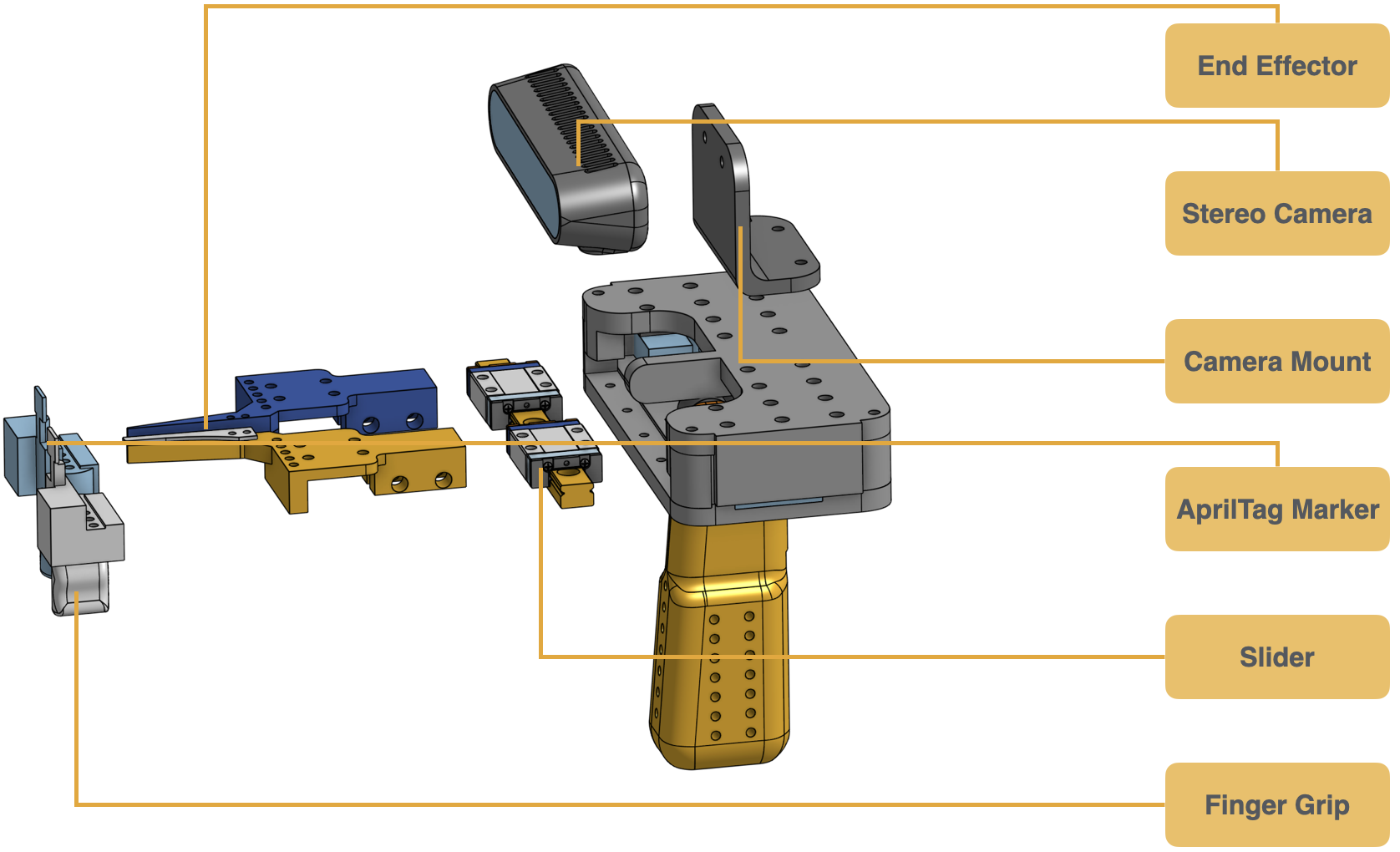}    
\caption{The handheld data collection device of SROI including gripper, finger slider, camera mount and hand holder} 
\label{fig:handheld_device_cad}
\end{center}
\end{figure}

\subsection{Hardware}
The hardware of the entire SROI system comprises three major components: the handheld device, the data collection unit, and the robot-side electric gripper.

The handheld data collection device is illustrated in Fig. \ref{fig:handheld_device_cad}. The end effector is designed to be long and thin, enabling it to easily navigate through the small gaps between clustered stems and leaves, effectively pushing obstacles aside to access the target stem. A cut-and-hold structure is employed in the gripper Fig. \ref{fig:handheld_device_cad} gripper, allowing it to cut stems using the blade while still retaining the stem in the gripper. For strawberry picking, the gripper can sever the stem just above the fruit and hold the strawberry until it is transported to the collection bin. For leaf or stolon removal, the gripper can cut the target object and discard it. The gripper can be replaced with other designs or functionalities. The finger slider (Fig. \ref{fig:handheld_device_cad} finger slider) allows the operator to easily manipulate the gripper using their fingers. Additionally, two AprilTags are mounted on the top side of the finger slider, enabling the detection of the gripper's opening in the camera view by tracking the position of the AprilTags. The gripper and finger slider are mounted on a sliding mechanism and hitched using two elastic bands, allowing the opening to be easily controlled with two fingers—closing when the fingers are pressed and opening when the finger pressure is released. The overall dimensions of the handheld data collection device are 125*60*27 (mm), excluding the grippers and camera mount, which makes it significantly smaller than the UMI device.

\begin{figure}
\begin{center}
\includegraphics[width=8.4cm]{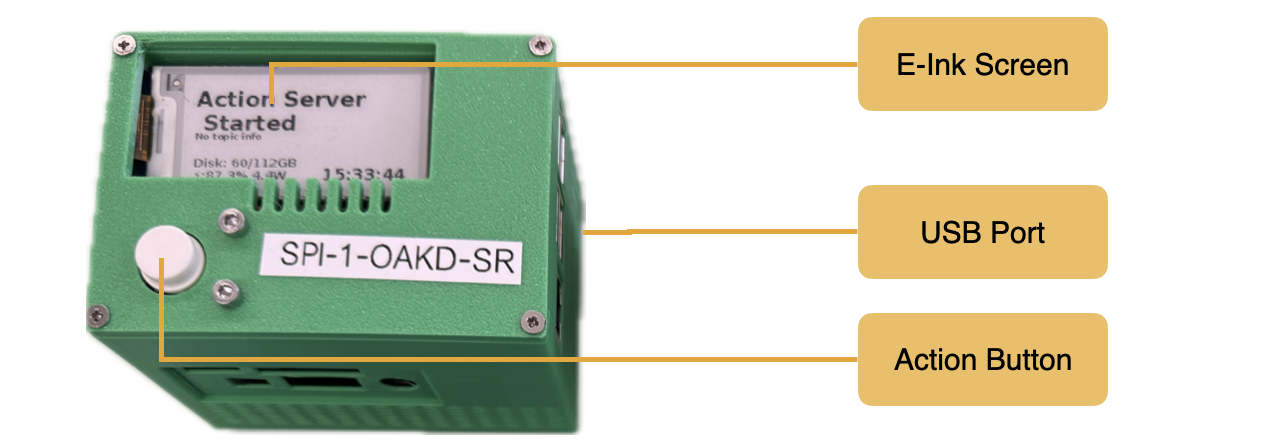}    
\caption{Data collection unit with E-ink screen, USB port for connecting to camera and a user control button} 
\label{fig:data_collection_unit}
\end{center}
\end{figure}

The data collection unit (Fig. \ref{fig:data_collection_unit}) serves to power the camera, operate the data collection system, and temporarily store data. The primary component of the unit is a Raspberry Pi 5 single-board computer (SBC), which drives the camera and captures all images and IMU data. A press button functions as a user control, allowing for the management of recording and the publishing of signals to segment demonstrations. An e-Link display is utilized to present system information, including recording status, disk availability, and battery information. An uninterruptible power supply (UPS) (please check the model and manufacturer) serves as the system's battery, providing 19.24 Wh of energy and allowing for approximately four hours of continuous recording (please verify). The entire data collection unit can be comfortably held in one hand while the other hand is used to collect manipulation data.


The robot-side electric gripper (Fig.~\ref{fig:sroi_collect_inference} right) shares the same appearance and camera mounting structure as the handheld device, enabling the direct application of policies trained on data collected from the handheld device to a robot arm equipped with the electric gripper. Two linear electric actuators are employed to drive the gripper and communicate with the robot controller via RS485.

\begin{figure}
\begin{center}
\includegraphics[width=8.4cm]{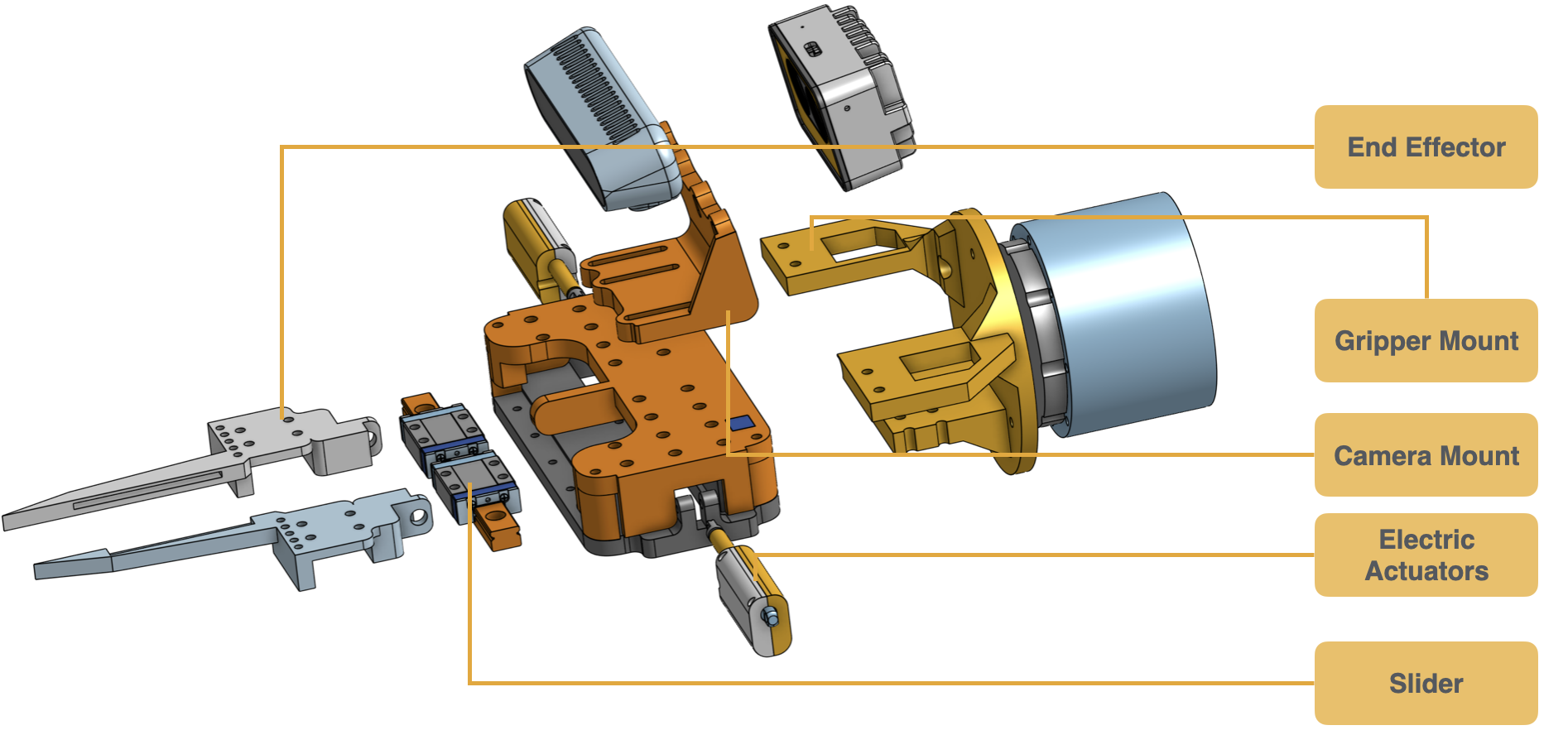}    
\caption{Left: E-gripper has two linear electric actuators with same appearance and camera mount as the handheld device } 
\label{fig:gripper}
\end{center}
\end{figure}

\subsection{Software}
The software of the SROI encompasses two components: the data acquisition system and the data post-processing system.  

\begin{figure}
\begin{center}
\includegraphics[width=8.4cm]{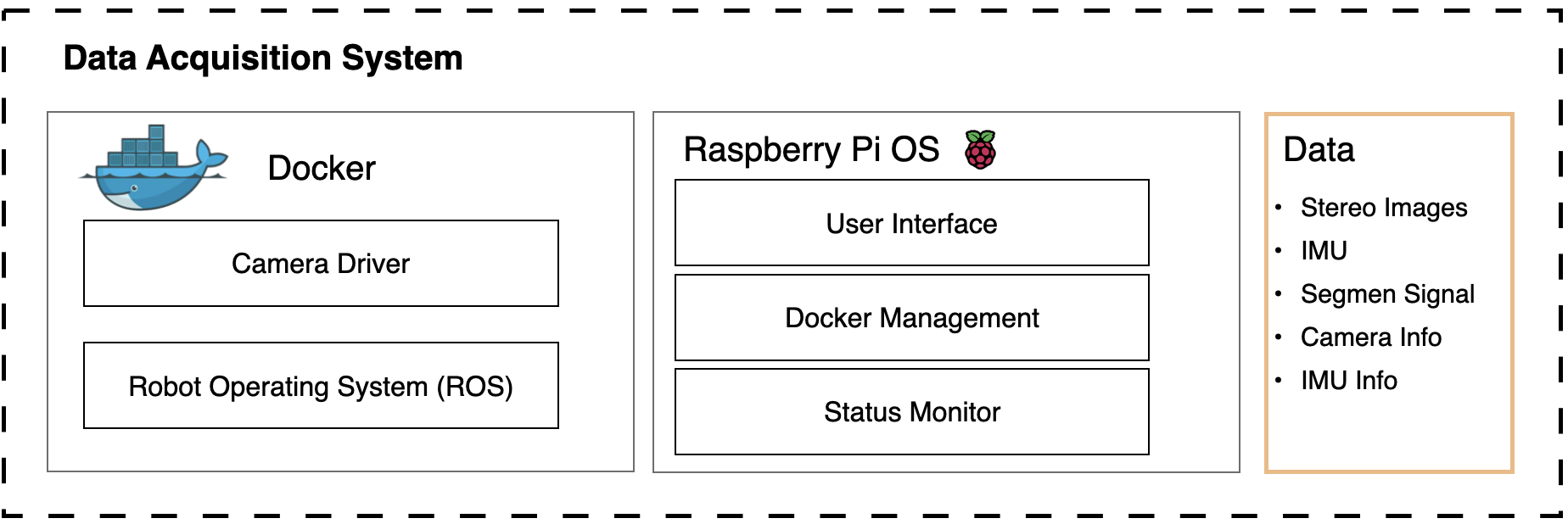}    
\caption{Data acquisition system on the data collection unit} 
\label{fig:data_acquisition_system}
\end{center}
\end{figure}

The data acquisition system, illustrated in Fig. \ref{fig:data_acquisition_system}, is deployed on a Raspberry Pi 5 within the data collection unit. This part of the system is responsible for driving the camera, timestamping the data, and recording the data into ROS bags.  The official Raspberry Pi OS serves as the main operating system. Within this environment, a Ubuntu 20.04 Docker container is deployed, which includes the Robot Operating System (ROS) and camera drivers. Additionally, auxiliary code monitors the system's status and displays information on the e-ink display.

\begin{figure}
\begin{center}
\includegraphics[width=8.4cm]{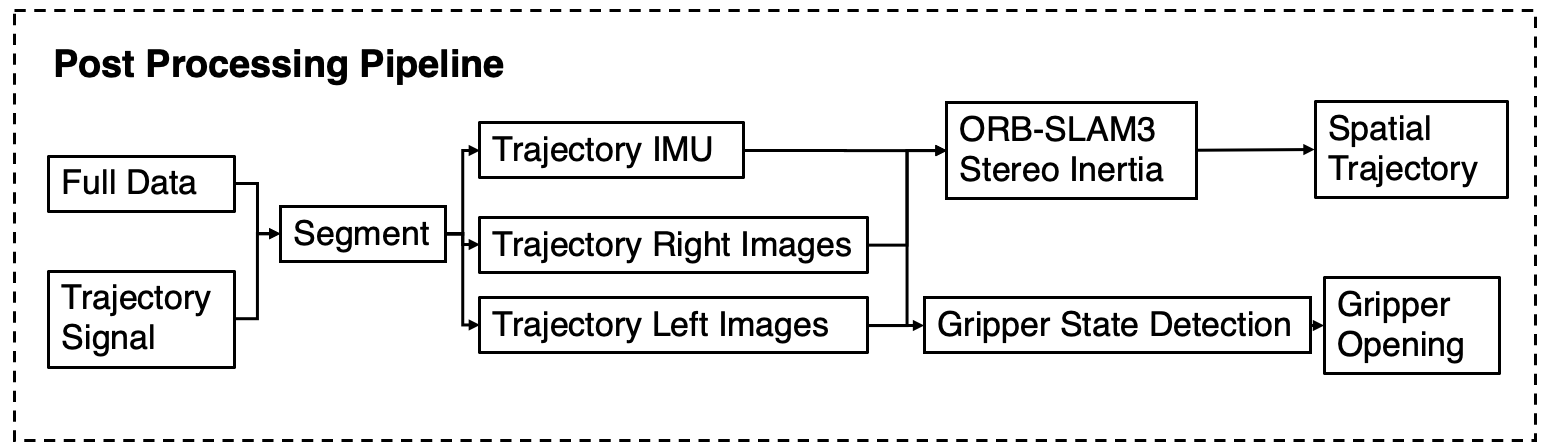}    
\caption{Data post processing pipeline} 
\label{fig:postprocessing_pipeline}
\end{center}
\end{figure}

\label{sec:post-processing}
The data post-processing system is depicted in Fig. \ref{fig:postprocessing_pipeline}. The primary objective of this component of the system is to process the recorded data bags and create datasets suitable for use with imitation algorithms, such as DP [ref] and ACT. In this post-processing pipeline, raw data is initially segmented into trajectories using the trajectory signal, ensuring that each trajectory contains only a single operation. Stereo image pairs, along with IMU information, are processed using the ORB-SLAM3 stereo inertia SLAM system to obtain the camera's spatial poses. Compared to the monocular inertia system used in UMI [ref], utilizing a stereo-based SLAM system significantly enhances pose estimation stability and eliminates the need for pre-mapping the scene. This adaptability is crucial for field data collection, as strawberry picking involves constant scene change. Theoretically, any stereo-based visual odometry or visual inertial odometry system could be employed in this context. Additionally, a gripper state detection system has been developed to recognize the April tags located on top of the gripper and estimate the gripper's opening.

\section{Experiments and Results}
This section outlines the specifics of clustered strawberry picking data collection utilizing the SROI device, encompassing camera setups and the data collection workflow. We also evaluated the spatial trajectory accuracy of the processed collected data.

\subsection{Camera setups}
We utilized two types of stereo cameras for data collection—the OAK-D-SR and the Intel Realsense D435i—to demonstrate the SROI's compatibility with multiple camera hardware configurations. Additionally, we anticipate that data from heterogeneous camera sources will aid in teaching imitation learning algorithms to develop a more generalized camera-agnostic manipulation policy. The configurations of each camera are detailed in table \ref{tb:camera_configs}.

\begin{table}[hb]
\begin{center}
\caption{Camera Configs}\label{tb:camera_configs}
\begin{tabular}{cccc}
\hline
Config & OAK-D-SR & Intel Realsense D435i  \\\hline
Resolution & 1280*720 & 848*480  \\ \hline
FPS & 30 & 30  \\ \hline
IMU(Hz) & 400 & 200(accel),250(gyro)  \\ \hline
Images & RGB (L,R) & Infrared (L,R), RGB  \\ \hline
\end{tabular}
\end{center}
\end{table}

Data collection occurred at two strawberry farms: one is a hill system field (Field 1) and the other is a vertical farm system (Field 2). Both strawberry farms are located in Hangzhou, Zhejiang, China, with the data collection period spanning from December 2024 to January 2025. Our primary goal was to capture the dexterous picking of clustered and occluded strawberries, which cannot be easily harvested using traditional direct pick methods. 


\subsection{Workflow}
The data collection workflow includes the following steps:
\begin{enumerate}
\item \textbf{Check Status and Start Recording}: In the field, once everything is set up, boot the data collection unit and verify all information to ensure all necessary data is streaming correctly. Long-press the button on the unit to begin recording.
\item \textbf{Identify Target Fruit:} Select the target strawberry for which you wish to collect picking data, typically strawberries that are occluded by stems, leaves, or other strawberries and present a challenge to be directly picked.
\item \textbf{Demonstration:} Use one hand (the action hand) to a position the SROI in front of the target fruit, where a robot arm could also feasibly position its end effector while maintaining visibility of the target fruit. With the other hand, hold the data collection unit and short-press the button to emit a demonstration start signal before the action hand moves. Following the start signal, use the SROI to approach, push away, or bypass obstacles, reaching the target fruit's picking point above the fruit to cut-hold the strawberry. Short-press the button again to emit a demonstration end signal.
\item \textbf{Repeat Steps 2-3:} Continue until sufficient data has been collected. In step 3, one can approach the picking point of the target strawberry without actually cutting the stem to preserve the strawberry, repeatedly reaching it from different starting points with various trajectories. This strategy allows for the collection of more usable data with a limited number of picked strawberries.
\item \textbf{Stop Recording:} Once enough data has been collected, long-press the control button to stop recording.
\item \textbf{Post Processing:} The recorded data can be post-processed as described in Section \ref{sec:post-processing} to extract each demonstration's end effector spatial movement trajectory and gripper action.
\end{enumerate}

\begin{figure}
\begin{center}
\includegraphics[width=8.4cm]{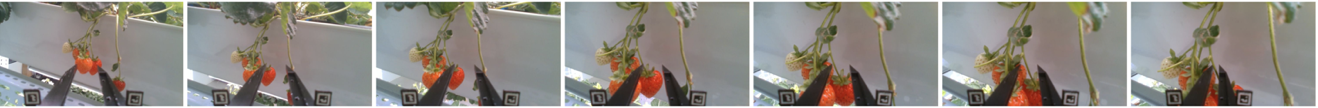}    
    \caption{Example of collected raw data} 
    \label{fig:picking_demostration_trajectory}
\end{center}
\end{figure}

Fig. \ref{fig:picking_demostration_trajectory} illustrates an example of raw data collected.

\subsection{Trajectory estimation accuracy evaluation}
To evaluate the accuracy of spatial trajectory pose estimation using the ORB-SLAM3 method, we mounted the SROI on a robot arm and collected data for pose estimation evaluation, including camera images and IMU data necessary for pose estimation, as well as the state of all the robot arm's joints to calculate the ground truth pose of the camera. Before data collection, hand-eye calibration was performed using the MoveIt hand-eye calibration package to obtain the camera pose on arm. Utilizing this transformation and the robot arm's URDF (Unified Robot Description Format), the ground truth pose of the camera was calculated from the states of all joints using forward kinematics. [Insert equation if needed].
The robot followed a script containing movements in each direction, 1 trajectories are collected in the lab facing ArUco makers [ref] to provide a reference when features are abundant, 7 evaluation data trajectories were collected in strawberry farm environments using the Intel Realsense D435i version of the SROI with stereo images for SLAM system without IMU. 


We used absolute pose error (APE) and relative pose error (RPE), where APE focus on the global trajectory consistency and RPE is a metric for evaluating local trajectory consistency. A package evo [reference] is used to evalute the consistency between the estimated trajectory and the referenced trajectory. 

The metrics of APE defined in equation~(\ref{eq:ape}), and RPE is defined in equation~(\ref{eq:rpe}) as follows:

\begin{equation} \label{eq:ape}
E_i = P_{\text{est},i} \ominus P_{\text{ref},i} = P_{\text{est},i}^{-1} P_{\text{ref},i}
\end{equation}

\begin{equation} \label{eq:rpe}
E_{i,j} = \delta_{\text{est}_{i,j}} \ominus \delta_{\text{ref}_{i,j}} = (P_{\text{ref},i}^{-1}P_{\text{ref},j})^{-1} (P_{\text{est},i}^{-1}P_{\text{est},j})
\end{equation}

\begin{description}
    \item[$E_i$] 
    The error term at the $i$-th instance. It represents the difference between the estimated pose and the reference pose.
    \item[$E_{i,j}$] 
    The relative pose error between the $i$-th and $j$-th instances. It represents the difference between the estimated and reference relative transformations.
    \item[$P_{\text{est},i}$] 
    The estimated pose at the $i$-th instance. 
    \item[$P_{\text{ref},i}$] 
    The reference pose at the $i$-th instance.
    \item[$\delta_{\text{est}_{i,j}}$] 
    The estimated relative pose change between the $i$-th and $j$-th instances.
    \item[$\delta_{\text{ref}_{i,j}}$] 
    The reference relative pose change between the $i$-th and $j$-th instances.

\end{description}

\begin{table*}[t]
    \centering
    \caption{APE and RPE Metrics for All Runs}
    \label{tab:ape_rpe_metrics}
    \begin{tabular}{l *{4}{S[table-format=1.5]} *{4}{S[table-format=1.5]}}
        \toprule
        & \multicolumn{4}{c}{APE} & \multicolumn{4}{c}{RPE} \\
        \cmidrule(lr){2-5} \cmidrule(lr){6-9}
        {Run} & {RMSE (m)} & {Mean (m)} & {Median (m)} & {Std (m)} & {RMSE (m)} & {Mean (m)} & {Median (m)} & {Std (m)} \\
        \midrule
        run1 (ArUco) & 0.00788 & 0.00604 & 0.00371 & 0.00506 & 0.00128 & 0.00092 & 0.00060 & 0.00089 \\
        run2         & 0.00736 & 0.00572 & 0.00385 & 0.00463 & 0.00109 & 0.00074 & 0.00047 & 0.00080 \\
        run3         & 0.00771 & 0.00603 & 0.00403 & 0.00481 & 0.00117 & 0.00079 & 0.00045 & 0.00086 \\
        run4         & 0.00738 & 0.00537 & 0.00298 & 0.00506 & 0.00117 & 0.00074 & 0.00041 & 0.00090 \\
        run5         & 0.01012 & 0.00755 & 0.00429 & 0.00674 & 0.00131 & 0.00083 & 0.00045 & 0.00101 \\
        run6         & 0.00748 & 0.00585 & 0.00398 & 0.00466 & 0.00118 & 0.00083 & 0.00052 & 0.00085 \\
        run7         & 0.00776 & 0.00593 & 0.00392 & 0.00501 & 0.00107 & 0.00070 & 0.00040 & 0.00081 \\
        run8         & 0.00751 & 0.00566 & 0.00346 & 0.00494 & 0.00135 & 0.00088 & 0.00052 & 0.00102 \\
        \midrule
        Average      & 0.00790 & 0.00602 & 0.00378 & 0.00511 & 0.00120 & 0.00080 & 0.00048 & 0.00089 \\
        \bottomrule
    \end{tabular}
\end{table*}

\begin{figure}
\begin{center}
\includegraphics[width=8.4cm]{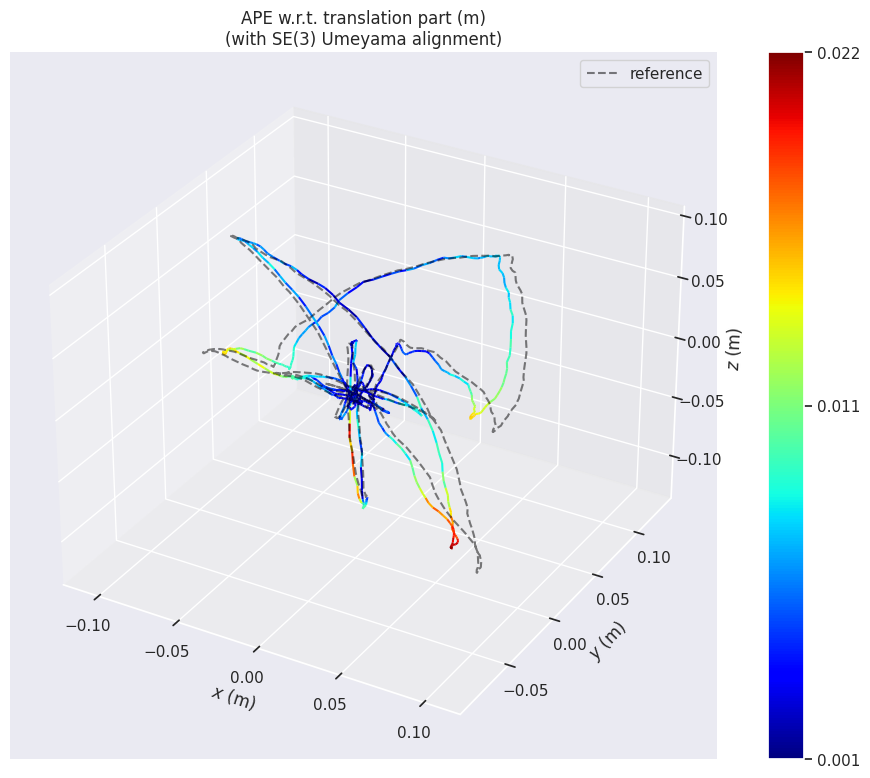}
    \caption{Spatial trajectory comparison between the SLAM estimated trajectory and the robot arm calculated reference trajectory. APE are annotated in color} 
    \label{fig:sample_traj_results_xyz}
\end{center}
\end{figure}

\begin{figure}[b]
    \centering
    
    \begin{subfigure}{0.45\textwidth}
        \centering
        \includegraphics[width=8.4cm]{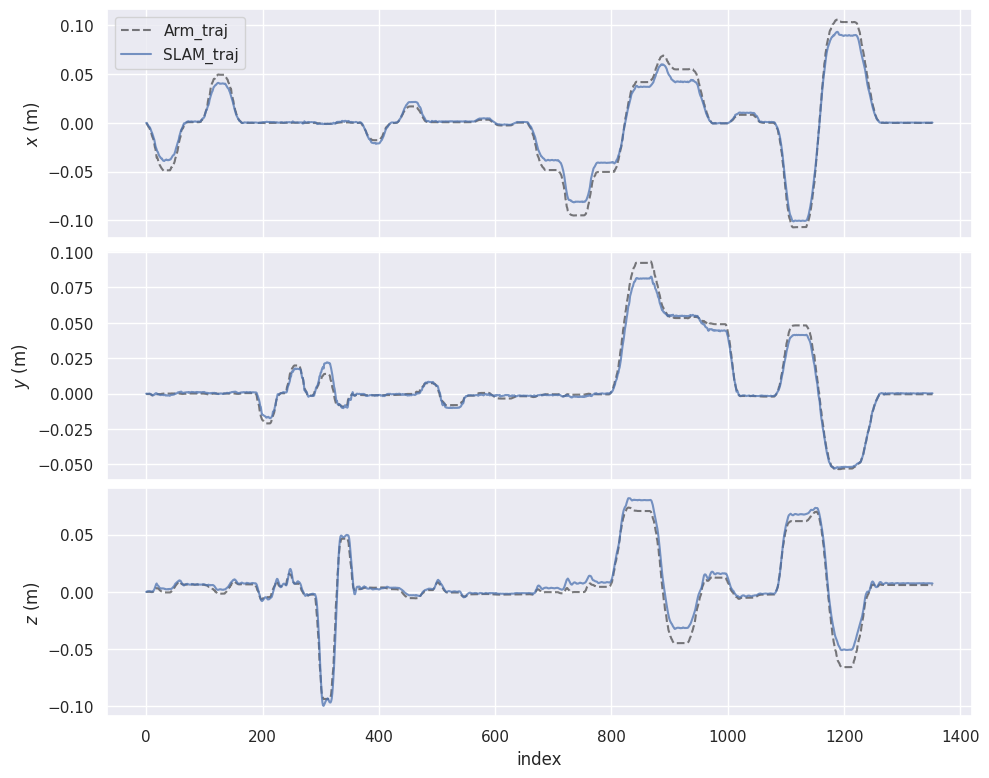}
        \caption{Comparison of translation part between the SLAM estimated trajectory and the robot arm calculated reference trajectory.}
        \label{fig:sample_traj_results_abc}
    \end{subfigure}
    
    \vspace{0.1cm} 
    
    \begin{subfigure}{0.45\textwidth}
        \centering
        \includegraphics[width=8.4cm]{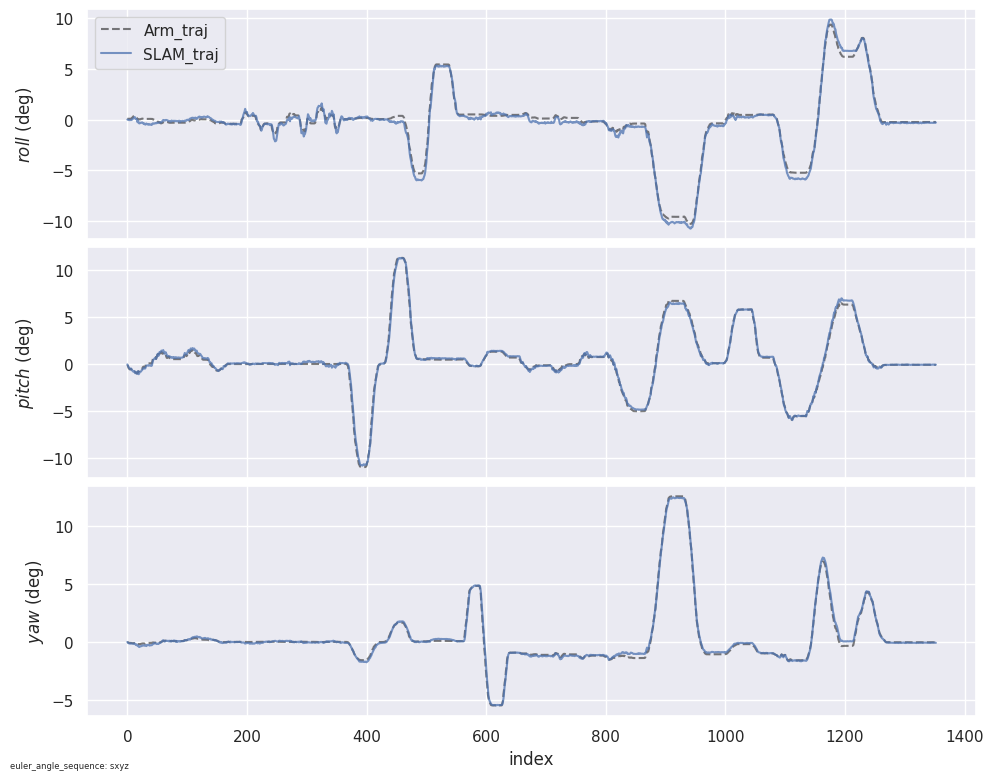}
        \caption{Comparison of rotation part between the SLAM estimated trajectory and the robot arm calculated reference trajectory.}
        \label{fig:sample_traj_results_def}
    \end{subfigure}
    
    \caption{Comparison a SLAM estimated trajectory and the robot arm calculated reference trajectory.}
    \label{fig:example_traj}
\end{figure}

A comparison between a SLAM-estimated trajectory and the robot arm's calculated reference trajectory is shown in Fig.~\ref{fig:sample_traj_results_xyz}. From Fig.~\ref{fig:sample_traj_results_abc} and Fig.~\ref{fig:sample_traj_results_def}, it can be observed that both the translation and rotation components of the estimated trajectory closely follow the reference trajectory. The results show that the RMSE of the APE translation component is 0.0075 m, which is smaller than the size of a strawberry fruit. Considering that the control policy operates the end effector in a closed-loop mode, errors can be corrected incrementally, making the actual effect of the APE even smaller than the reported value.

The results of all eight trajectories are reported in table~\ref{tab:ape_rpe_metrics}. As can be observed, the accuracy remains quite consistent across different runs and is also comparable to that obtained when facing ArUco markers (as seen in run 1). This indicates that the strawberry picking environment provides ample visual features for tracking, thus enabling the visual-based SLAM to achieve high accuracy. Consequently, this high level of precision contributes to generating reliable trajectories for imitation learning.

\section{Conclusion}

In this work, we introduced the SROI, a novel open-source device specifically designed for collecting dexterous manipulation data in strawberry farming. The SROI addresses the challenges associated with gathering data on dexterous strawberry field manipulations, such as picking occluded strawberries, by providing a portable, low-cost, and modular system for data collection. Our evaluation demonstrated the system's high precision in trajectory estimation, making it suitable for training imitation learning algorithms. Furthermore, we released an open-source dataset of strawberry picking demonstrations, which we believe will accelerate advancements in robotic manipulation for agriculture. The SROI represents a key step forward in automating complex farming tasks and reducing labor dependency. Future work will concentrate on deploying the SROI in larger-scale field trials to collect a substantial amount of demonstration data and to integrate it with advanced imitation learning algorithms. This will demonstrate the capability of a data-driven approach in solving dexterous robotic operations in strawberry farming.

\begin{ack}
This work is supported by the Natural Science Foundation of Zhejiang Province, China (Grant No. LD24C130003). We would like to express our gratitude to the strawberry grower in Shunba Village for providing a field for data collection for this research
\end{ack}

\bibliographystyle{agsm} 
\bibliography{ifacconf}             
                                                   








\end{document}